%% file: iclr2025_conference.tex
\documentclass{article}

\usepackage{iclr2025_conference,times}

\input{math_commands.tex}

\usepackage{hyperref}
\usepackage{url}

\usepackage{cleveref}
\usepackage{booktabs}
\usepackage{amsmath}
\usepackage{graphicx}
\usepackage{tikz}
\usepackage{makecell}
\usepackage{enumitem}
\usepackage{color, colortbl}
\usepackage{tikz}
\usepackage{subcaption}
\usepackage{wrapfig}
\usepackage{multirow}
\usepackage{listings}
\usepackage{algorithm}
\usepackage{algorithmic}
\usepackage{stfloats}
\usepackage{xspace}
\usepackage{booktabs}
\usepackage{xcolor}
\usepackage{afterpage}
\usepackage{setspace}
\usepackage{tocloft}
\usepackage{appendix}

\definecolor{tabhighlight}{HTML}{e5e5e5}
\definecolor{t2i}{HTML}{F0D695}
\definecolor{control}{HTML}{C9B1D3}
\definecolor{semantic}{HTML}{D26F70}
\definecolor{element}{HTML}{98B567}
\definecolor{repaint}{HTML}{82ACD1}

\crefname{section}{Sec.}{Secs.}
\Crefname{section}{Section}{Sections}
\crefname{table}{Tab.}{Tabs.}
\Crefname{table}{Table}{Tables}
\crefname{figure}{Fig.}{Figs.}
\Crefname{figure}{Figure}{Figures}
\crefname{equation}{Eq.}{Eqs.}
\Crefname{equation}{Equation}{Equations}
\crefname{algorithm}{Alg.}{Algs.}
\Crefname{algorithm}{Algorithm}{Algorithms}

\newcommand{\basemodelnc}{\text{FLUX.1-dev}\xspace}
\newcommand{\basemodel}{\text{FLUX.1-dev} \cite{flux}\xspace}
\newcommand{\fillmodelnc}{\text{FLUX.1-Fill-dev}\xspace}
\newcommand{\fillmodel}{\text{FLUX.1-Fill-dev} \cite{flux}\xspace}

\title{
ACE++:Instruction-Based Image Creation and Editing via Context-Aware Content Filling
}

\author{%
\xspace \xspace \xspace \quad \xspace\xspace \quad  
Chaojie Mao \quad 
Jingfeng Zhang \quad
Yulin Pan \quad
Zeyinzi Jiang \quad
Zhen Han \quad
\\[3pt]
\xspace \xspace \quad \xspace \quad \xspace \xspace \quad
\xspace \quad \xspace \quad \xspace \quad 
\xspace \quad \xspace \quad \xspace \quad \xspace \quad \xspace \quad
\textbf{Yu Liu} \quad
\textbf{Jingren Zhou} \quad 
\\[10pt]
\xspace \quad \xspace \quad
\xspace \quad \xspace \quad \xspace \quad
\xspace \quad \xspace \quad \xspace \quad 
\xspace \quad \xspace \quad \xspace \quad 
\xspace \xspace \xspace \quad \xspace \quad
Tongyi Lab \quad 
}

\iclrfinalcopy 

\begin{document}

\maketitle

\begin{abstract}
\input{sections/0.abstract}
Code and models will be available on the project page: \url{https://ali-vilab.github.io/ACE_plus_page/}.
\end{abstract}

\input{sections/1.introduction}
\input{sections/2.related_work}
\input{sections/3.method}

\input{sections/4.experiments}
\input{sections/5.conclusion}

\clearpage
\bibliography{iclr2025_conference}
\bibliographystyle{iclr2025_conference}


\end{document}

%% file: math_commands.tex
\usepackage{amsmath,amsfonts,bm}

\def\eqref#1{equation~\ref{#1}}

\def\1{\bm{1}}

\DeclareMathAlphabet{\mathsfit}{\encodingdefault}{\sfdefault}{m}{sl}
\SetMathAlphabet{\mathsfit}{bold}{\encodingdefault}{\sfdefault}{bx}{n}



%% file: sections/0.abstract.tex


We report ACE++, an instruction-based diffusion framework that tackles various image generation and editing tasks. 
Inspired by the input format for the inpainting task proposed by \fillmodelnc, we improve the Long-context Condition Unit (LCU) introduced in ACE and extend this input paradigm to any editing and generation tasks. To take full advantage of image generative priors, we develop a two-stage training scheme to minimize the efforts of finetuning powerful text-to-image diffusion models like \basemodelnc. In the first stage, we pre-train the model using task data with the 0-ref tasks from the text-to-image model. There are many models in the community based on the post-training of text-to-image foundational models that meet this training paradigm of the first stage. For example, \fillmodelnc deals primarily with painting tasks and can be used as an initialization to accelerate the training process. In the second stage, we finetune the above model to support the general instructions using all tasks defined in ACE. To promote the widespread application of ACE++ in different scenarios, we provide a comprehensive set of models that cover both full finetuning and lightweight finetuning, while considering general applicability and applicability in vertical scenarios. The qualitative analysis showcases the superiority of ACE++ in terms of generating image quality and prompt following ability.

%
%
%
%
%

%

%% file: sections/1.introduction.tex
\section{Introduction}\label{sec:intro}

Recently, the field of visual generation has made significant progress due to breakthroughs in diffusion models. 
The open-source of foundational text-to-image models such as \basemodel and Stable Diffusion XL has facilitated the development of many image editing methods tailored for specific applications.
However, the development of universal image editing models is lagging behind. While some literature focuses on training versatile image creators who excel at generating images based on user instructions, the aesthetics of the generated images are far from satisfactory.
To fully leverage the capabilities of image generative priors, it is customary to finetune text-to-image foundational models rather than train them from scratch. However, due to the input-output format discrepancies between text-to-image generation tasks and other image editing tasks, the tuning process often converges slowly. 


%

In the community, many editing or reference generation methods are based on post-training of text-to-image foundational models. Leveraging the powerful generative capabilities of these models, task adaptation can be achieved with relatively small data scales and lower training costs. For instance, \fillmodel achieves inpainting task training by channel-wise concatenating the image to be edited, the area mask, and the noisy latent on the basis of the text-to-image model. Inspired by this, we modified the input paradigm Long-context Condition Unit (LCU) introduced in ACE ~\cite{ace} to reduce the substantial model adaptation costs brought by the introduction of multimodal inputs. Specifically, for the 0-ref tasks where the inputs do not include a reference image such as controllable generation, inpainting, and single-image editing, ACE ~\cite{ace}'s LCU concatenates inputs in the token sequence dimension for the model. Compared to text-to-image generation, this approach introduces an additional conditional sequence into the attention input of the diffusion transformer, resulting in increased computational load during the attention phase and thereby increasing the training costs for model adaptation. By changing the conditional input for these tasks from sequence concatenation to channel dimension concatenation, we can effectively reduce the model adaptation costs. Furthermore, we propose an improved LCU input paradigm called LCU++ by extending this paradigm to arbitrary editing and reference generation tasks.

Based on the input definition of LCU++, we divide the entire training process into two stages. The first stage involves pre-training the model using task data with 0-ref tasks from the text-to-image model. There are many models in the community based on the post-training of text-to-image foundational models that align with the training paradigm of the first stage. For instance, \fillmodel mainly focuses on inpainting tasks and can be used as the model initialization to accelerate the training process. In the second stage, we fine-tune the above model to support the general instructions using all data collected in ACE ~\cite{ace}. During the training phase, we simultaneously focus on the model's ability to reconstruct the input reference images and generate the target images to guide the model in learning context-aware information. In the inference phase, the model fills in the content according to the mask area to be generated, enabling it to be compatible with both local editing and global regeneration tasks simultaneously.

The eight categories of generation tasks inducted by ACE ~\cite{ace} have a wide range of application scenarios. In addition to continuing the all-round editing and generation model trained by the ACE ~\cite{ace} dataset, ACE++ provides a range of lightweight models for several of the most widely used scenarios in the community, such as portrait consistency, subject consistency, local editing, and repainting in vertical scenarios. This aims to facilitate the open-source community in exploring more innovative scenarios. We give a qualitative analysis to showcase the performance of ACE++.

%% file: sections/2.related_work.tex
\section{Related Works}
\textbf{Multimodal-guided Image Generation and Editing.}
Large-scale text-to-image models~\cite{glide,imagen,dalle2,ldm,sdxl,dalle3,midjourney,wanx,ernie_vilg,pixart,sd3,kolors,hunyuan_dit,flux} achieve remarkable image fidelity; however, they are less suitable for image editing tasks. Furthermore, the integration of multimodal inputs is essential for expanding their range of applications.
To add spatial controls to original text-to-image models, controllable generation methods~\cite{controlnet, scedit} incorporate extra low-level visual features as input. 
Based on an input source image, instruction-based editing methods~\cite{ip2p,magicbrush,emuedit,stylebooth} are trained to follow textual or multimodal editing instructions.
Region-based editing methods only make changes in the area specified by input mask, guided by input prompt~\cite{brushnet,powerpaint} or another subject image~\cite{anydoor,largen}.
Guided by an exemplar image, reference-based generation methods aims to generate a new image with high fidelity of stylistic~\cite{styledrop} or facial~\cite{instantid} features.

\textbf{Unified Image Generation and Editing Framework.}
Inspired by the multi-task processing capability of VL-LLMs with multimodal understanding, recent works have explored unified frameworks for multi-task visual generation. OmniGen~\cite{omnigen} uses a language model as an initialization and jointly models text and images within a single model to achieve unified representations across different modalities. ACE~\cite{ace} constructs a unified input paradigm, LCU, for multimodal tasks within the Diffusion Transformer framework, allowing for training multimodal generative models compatible with multi-task data. UniReal~\cite{unireal} treats image-level tasks as discontinuous video generation, concatenating multiple input conditions along the sequence dimension to train a basic full-attention model. These methods face the significant training cost associated with training models from scratch. OminiControl~\cite{ominicontrol} employs a parameter reuse mechanism to inject input conditions by reusing the model's basic parameters. However, this approach limits support for multitasking.

%% file: sections/3.method.tex
\section{Method}\label{sec:datas}

We present LCU++, an innovative universal input paradigm designed to be compatible with a wide range of tasks, facilitating the complementary integration of knowledge across different tasks. 
In addition, we design a two-stage training scheme to minimize the efforts of fitting the LCU++ paradigm, enabling highly efficient fine-tuning of the pretrained text-to-image models.
We will describe them in detail below.

\begin{figure*}[ht]
    \scriptsize
    \centering
    \includegraphics[width=1.0\linewidth]{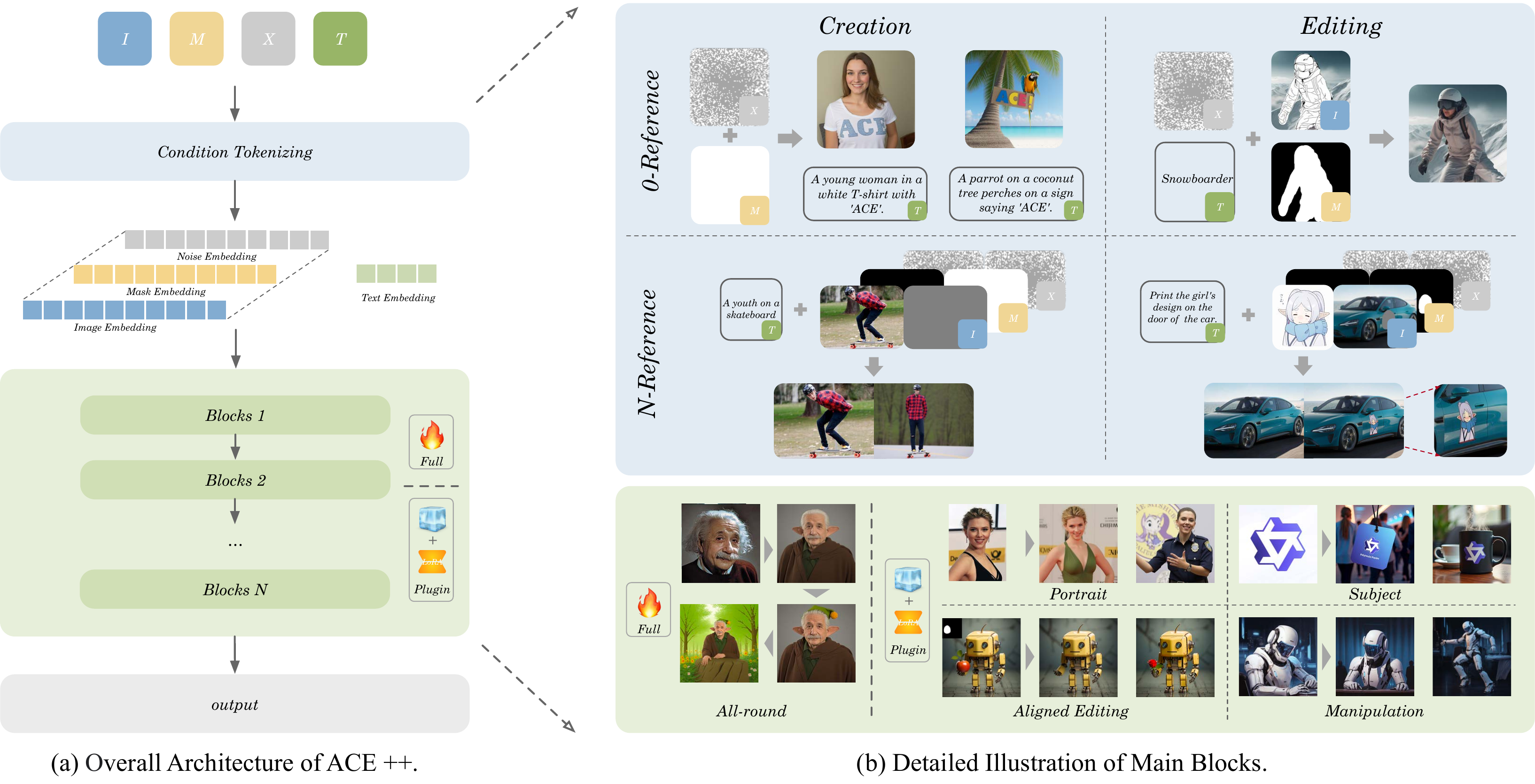}
    \caption{\textbf{The illustration of ACE++ framework.} The input paradigm of LCU++ is tokenized according to the respective methods of the editor and the creator, achieving different functional suites through full or partial fine-tuning.}
    \label{fig:method}
\end{figure*}

\begin{figure*}[ht]
    \scriptsize
    \centering
    \includegraphics[width=1.0\linewidth]{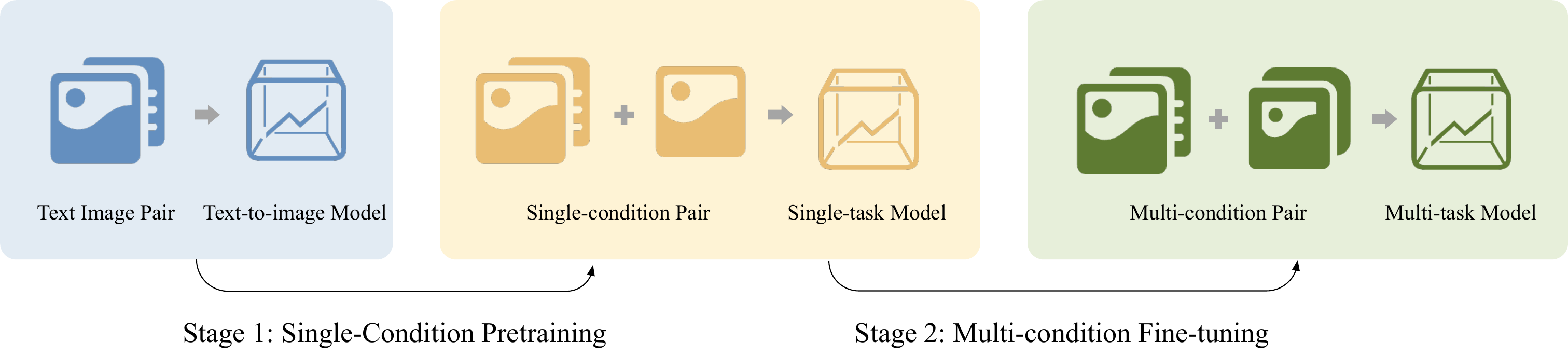}
    \caption{\textbf{The illustration of two-stage training scheme.}}
    \label{fig:stage}
\end{figure*}

\subsection{Preliminary}

We first review the Long-context Condition Unit (LCU), a universal multimodal input paradigm introduced in ACE ~\cite{ace}. The LCU is constructed by flattening and combining several Conditional Units (CUs) to support any number of input condition units (CU). CU consists of a textual instruction $T$ that describes the generation requirements, along with visual information $V$. Here, $V$ includes a set of images representation $I$, which can be defined as $I=\O$ (if there are no source images) or $I = {I^1, I^2, ..., I^N}$ (if there are source images), and corresponding masks $M = {M^1, M^2, ..., M^N}$. Unless otherwise specified, $I$ and $T$ refer to the feature representations of images and instructions, respectively. In addition, a noise unit is utilized to drive the generation process, which is constructed with the noisy latent $X_{t}$ at timestep $t$ and mask $M^{N}$. The overall formulation of the CU and the LCU is as follows:

\begin{equation}
    \text{CU} = \{T, V\}, \quad V=\{[I^{1}; M^{1}], [I^{2}; M^{2}], \dots, [I^{N}; M^{N}], [X_{t}; M^{N}]\},
\end{equation}

\begin{equation}
    \text{LCU} = \{\{T_{1}, T_{2}, \dots, T_{m}\}, \{V_{1}, V_{2}, \dots, V_{m}\}\}
\end{equation}

Considering the generation tasks without reference image (denoted as 0-Ref tasks), the corresponding input format $\text{LCU}_{\text{0-ref}}$ can be expressed as follows:

\begin{equation}
    \text{LCU}_{\text{0-ref}} = \{\{T\}, \{V\}\} , \quad V_{\text{0-ref}}=\{[I^{in}; M^{in}], [X_{t}; M^{in}]\}
\end{equation}

where $I^{in}$ and $M^{in}$ is optional and represent the input image and mask.


\subsection{Improved Long-context Condition Unit}
In the situation of existing input image and mask, usually for spatial aligned editing tasks, ACE ~\cite{ace} appends the conditional sequence $[I^{in}; M^{in}]$ to the noise sequence $[X_{t}; M^{in}]$, which disrupts the established context perception framework honed during text-to-image generation training. This inefficiency necessitates prolonged training periods, which is suboptimal for performance. Rather than sequence concatenation, We concatenate the conditional context and noise context in channel dimension, thereby mitigating the context perception disruption. The new format can be represented as:

\begin{equation}
    \text{LCU}^{++}_{\text{0-ref}} = \{\{T\}, \{V^{++}\}\} , \quad V^{++}_{\text{0-ref}}=\{[I^{in}; M^{in}; X_{t}]\}
\end{equation}

This input format of 0-ref tasks can then be extended to $N$-ref tasks by sequentially concatenating multiple reference context ${V^{++}}$'s along with the target context. Thus, the overall LCU++ paradigm can be formulated as:

\begin{equation}
\begin{split}
    \text{LCU}^{++} & = \{\{T_{1}, T_{2}, \dots, T_{m}\}, \{V^{++}_{1}, V^{++}_{2}, \dots, V^{++}_{m}\}\} \\
    V^{++} &=\{I^{1}; M^{1}; X^{1}_{t}], [I^{2}; M^{2};X^{2}_{t}], \dots, [I^{N}; M^{N}; X^{N}_{t}]\}
\end{split}
\end{equation}

where $X^{i}_{t}, i \in \{1, \ldots, N\}$ represents the $i$th noisy latent at timestep $t$.

\subsection{Model Architecture}

We illustrate the overall model architecture in ~\cref{fig:method}, which integrates the aforementioned LCU++ paradigm into the full attention framework of \basemodel. In each condition unit, the input image, mask, and noise are concatenated along the channel dimension to form the CU feature map. These CU feature maps are mapped to sequential tokens through the x-embed layer. Next, all CU tokens are concatenated to serve as the inputs of the transformer layer.

We design a novel and effective objective to optimize the training process. Given the input images $I^{i}, i\in \{1, 2, \dots, N\}$ and the target output image $I^{o}$, the noisy latent space $\textbf{x}_t = \{X^{i}_{t}\}, i \in \{1, \ldots, N\}$ can be constructed using a linear interpolation method from $\{I^{1}, I^{2}, \ldots, I^{N-1}, I^{o}\}$. Here, $I^{N}$ represents the sample that needs to be modified by the model. For editing tasks, this sample is the one to be edited. For reference generation tasks, this sample is an all-zero sample. The model is trained to predict the velocity $\textbf{u}_t = d\textbf{x}_t/d_t$, guiding the sample $\textbf{x}_t$ towards the sample $\textbf{x}_1$. The training objective is to minimize the mean squared error between the predicted velocity $\textbf{v}_t$ and the ground truth velocity $\textbf{u}_t$, expressed as the loss function:

\begin{equation}
\begin{split}
    \mathcal{L} &= \mathbb{E}_{t, \mathbf{x}_0, \mathbf{x}_1} \left\| \mathbf{v}_t - \mathbf{u}_t \right\|^2. \\
                &= \sum^{i=0}_{N-1}\mathbb{E}_{t, \mathbf{x}_0, \mathbf{x}_1} \left\| \mathbf{v}^{i}_t - \mathbf{u}^{i}_t \right\|^2 + \mathbb{E}_{t, \mathbf{x}_0, \mathbf{x}_1} \left\| \mathbf{v}^{N}_t - \mathbf{u}^{N}_t \right\|^2 \\
                &= \mathcal{L}_{\text{ref}} + \mathcal{L}_{\text{tar}}
\end{split}
\end{equation}
where $\mathcal{L}_{\text{ref}}$ represents the reconstruct loss of the first N-1 reference samples and is equal to 0 in 0-ref tasks. $\mathcal{L}_{\text{tar}}$ represents the loss of the target sample being predicted. In this way, the model possesses context-aware generation capabilities.

\subsection{Two-Stage Training Scheme}

We propose a two-stage training scheme as shown in ~\cref{fig:stage}. In the first stage, we train the model with 0-ref tasks, builded on a text-to-image model. This approach leverages the foundational generation capabilities of the text-to-image model without introducing additional sequences, allowing the model to quickly develop support for conditional inputs. In the second stage, we finetune the model with all 0-ref and $N$-ref tasks, to enable support for general instructions.

\subsection{A Suite of Toolkits}

As previously mentioned, ACE++ enhances the input paradigm of LCU to leverage the generative capabilities of the underlying model effectively. Using the \basemodel, we fully finetune a composite model with ACE's data to support various editing and reference generation tasks through an instructive approach. Additionally, for widely used areas such as portrait preservation, subject-driven generation, localized editing, and image variation, we train lightweight and domain-stable fine-tuned models using lightweight fine-tuning strategies like LoRA. Consequently, ACE++ provides a comprehensive toolkit for image editing and generation to support various applications.

%% file: sections/4.experiments.tex
\input{figures/subject/subject}
\input{figures/portrait/face}
\input{figures/local/local}
\input{figures/general/general}
\input{figures/local_ref/local_ref}
\section{Experiments}\label{sec:exp}
\subsection{Implementation Details}
We train our models based on various task data collected from ACE ~\cite{ace}. According to a two-stage training process, we use \basemodel as the base model, which is widely used by the community due to its superior generative capabilities. During the training phase, we first pre-train this text-to-image model on 0-ref tasks data. Then, we continue with a full finetuning process on the full dataset using this pre-trained model. For task-specific models in four vertical domains, we consider that the \fillmodel model, which is an inpainting model trained on \basemodel, aligns with the 0-ref tasks training method of the first stage. Therefore, these models can be directly lightweight finetuned on the basis of \fillmodel.

We use AdamW~\citep{adamw} optimizer to train the model with the weight decay as 1e-2 and the learning rate as 1e-3. We employ gradient clipping with a threshold of 1.0 using the L2 norm to stabilize the training process. Since the \basemodel model is trained using guidance distillation for different guidance values in classifier-free guidance, it is important to consider that the training process involves a mix of training conditional and unconditional models. Therefore, an essential training technique is to set the guidance scale $\omega$ to 1.0 and the unconditional probability to 0.1 considering the following analysis:

\begin{equation}
\begin{split}
    & \mathbf{v}_t = \mathbf{v}_t(\O) + \omega (\mathbf{v}_t(\textbf{c}) - \mathbf{v}_t(\O)) \\
    & so \quad \mathbf{v}_t = \mathbf{v}_t(\textbf{c}) \quad (\omega = 1.0)
\end{split}
\end{equation}

where $\mathbf{v}_t$ is the model's prediction with classifier-free guidance. $\mathbf{v}_t(\O)$ and $\mathbf{v}_t(\textbf{c})$ are the unconditional model prediction and conditional model prediction, respectively.

\subsection{Qualitative Results}

In this section, we present the visualization results of ACE++, particularly focusing on three common types of editing and generation as following:

\textbf{Reference Generation} is used to generate images with consistent identity for specific objects in the given reference image. As shown in ~\cref{fig:subject}, ACE++ is used in the subject-driven image generation task to maintain the consistency of a specific subject in different scenes. We can observe that, by maintaining the subject's consistency, it can be applied to various interesting application scenarios. The result in portrait-consistency generation scenarios is shown in ~\cref{fig:portrait}. Portrait-consistent generation is an important generative task that can be applied in film special effects, advertising production, and e-commerce design. It can be seen that ACE++ possesses a high-quality portrait image generation capability. 

\textbf{Local Editing and General Editing} is a crucial tool for secondary image refinement, which is demonstrated in ~\cref{fig:local}. Traditional inpainting methods struggle to redraw images while maintaining the original structural information of the edited area. ACE ~\cite{ace} has constructed a large amount of data for the detailed editing tasks, allowing the diverse and refined editing capabilities based on these data. ACE++ can also handle many open-ended editing and generation capabilities, such as perspective changes, color changes, and scene changes, as depicted in ~\cref{fig:general}. By using flexible commands or descriptions, we encourage developers to explore more interesting application scenarios with ACE++.

\textbf{Local Reference Editing} involves reference-guided generation for specific regions of an image, offering a wide range of applications such as try-on and product image generation. Leveraging ACE++, these tasks can be supported in a zero-shot manner without training these tasks in the training phase. ~\cref{fig:local_ref} illustrates the effectiveness of the application in relevant scenarios.

%% file: figures/subject/subject.tex
\begin{figure*}[t]
    \scriptsize
    \centering
    \includegraphics[width=\linewidth]{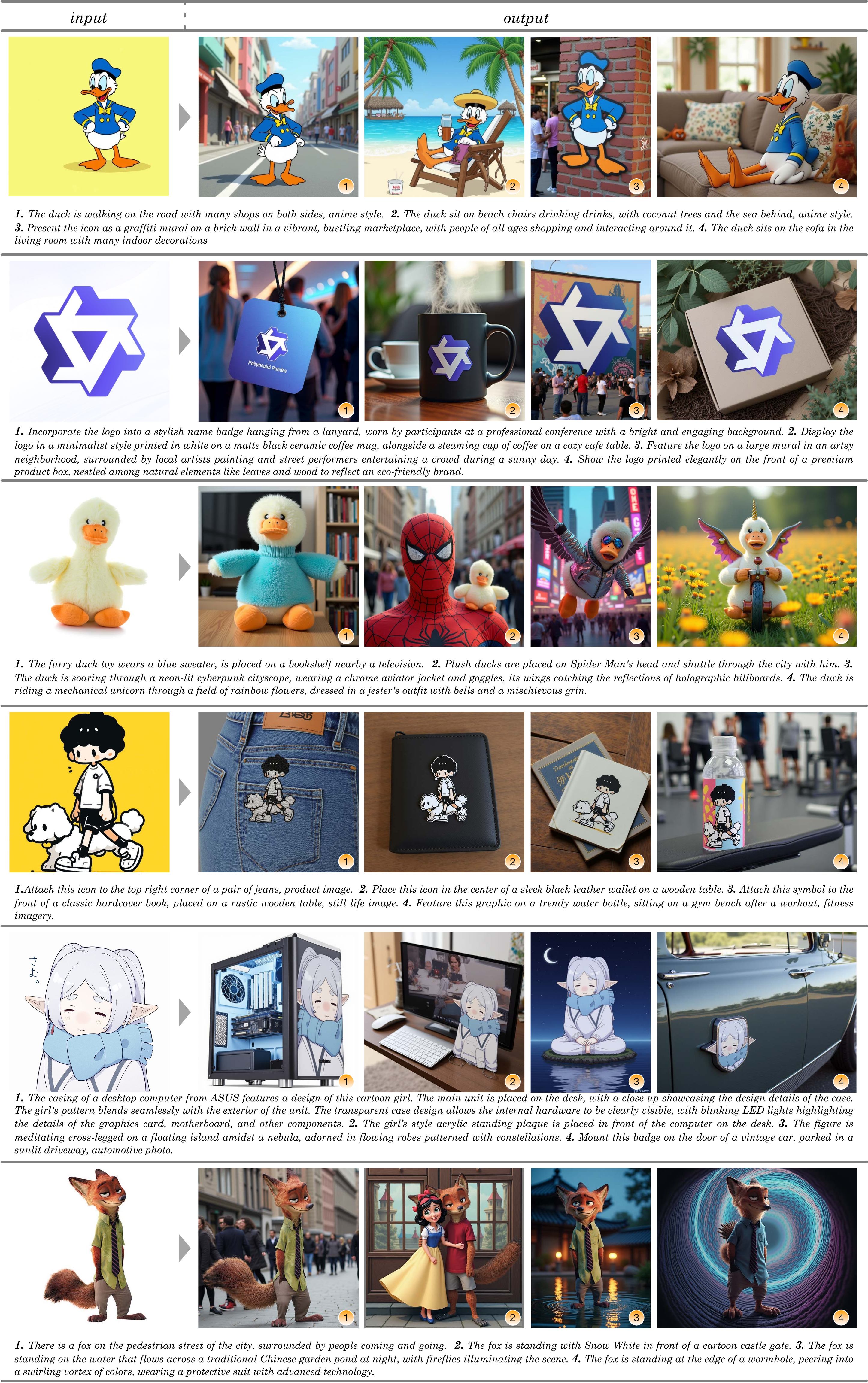}
    \caption{The visualization of subject-driven generation.}
    \vspace{-15pt}
    \label{fig:subject}
\end{figure*}

%% file: figures/portrait/face.tex
\begin{figure*}[t]
    \scriptsize
    \centering
    \includegraphics[width=\linewidth]{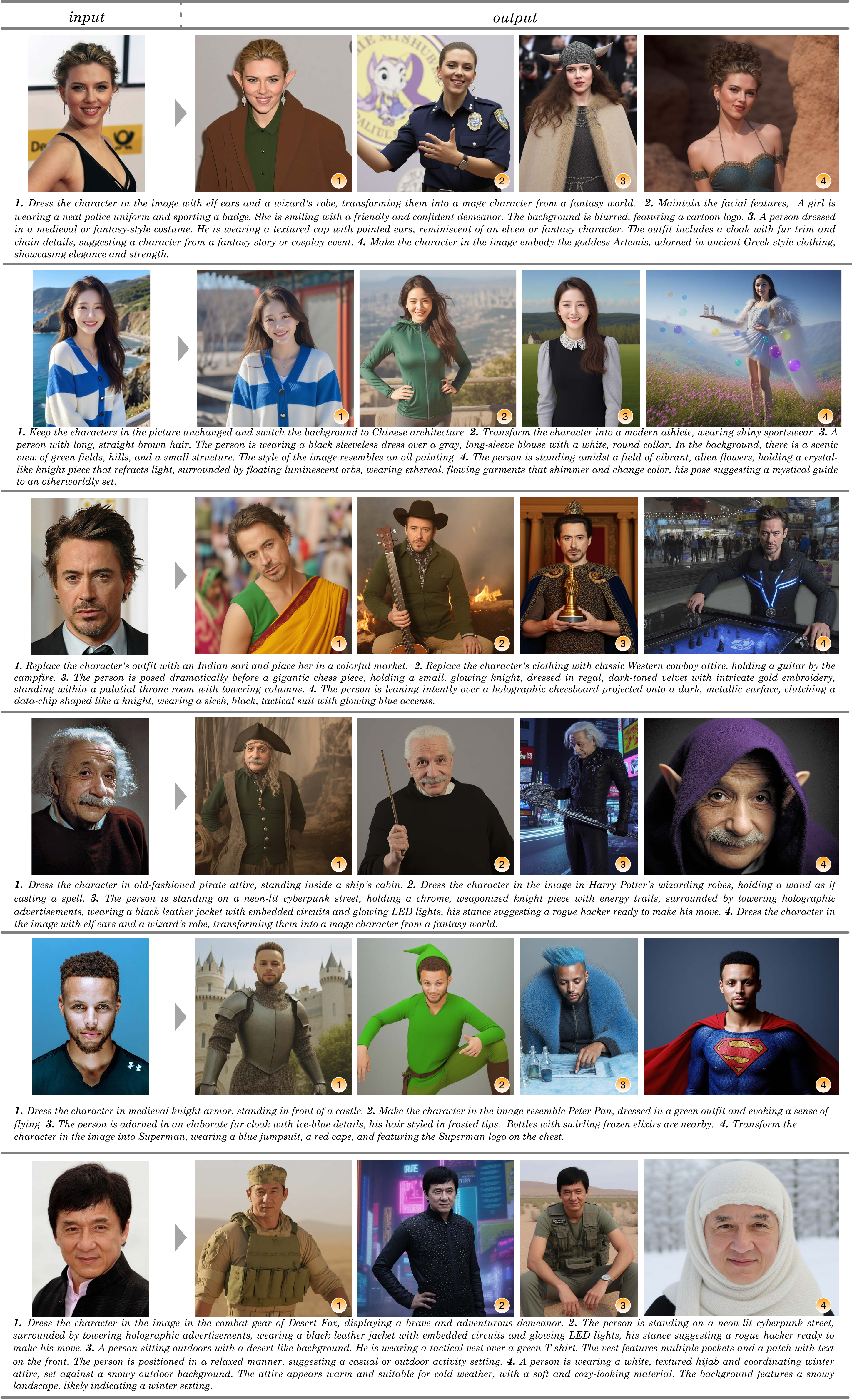}
    \caption{The visualization of portrait-consistency generation.}
    \vspace{-15pt}
    \label{fig:portrait}
\end{figure*}

%% file: figures/local/local.tex
\begin{figure*}[t]
    \scriptsize
    \centering
    \scalebox{0.96}{\includegraphics[width=\linewidth]{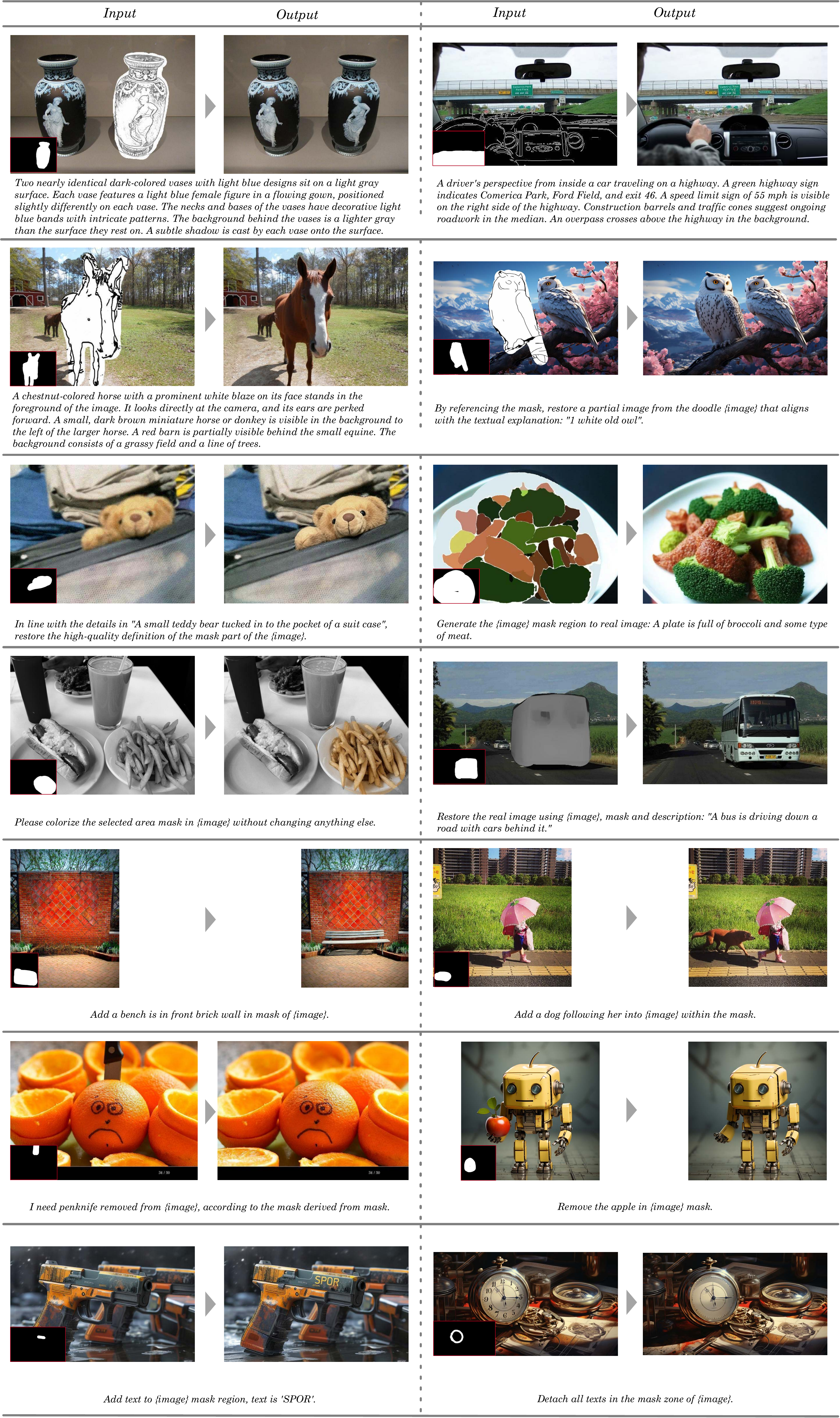}}
    \caption{The visualization of local editing.}
    \vspace{-15pt}
    \label{fig:local}
\end{figure*}

%% file: figures/general/general.tex
\begin{figure*}[t]
    \scriptsize
    \centering
    \includegraphics[width=\linewidth]{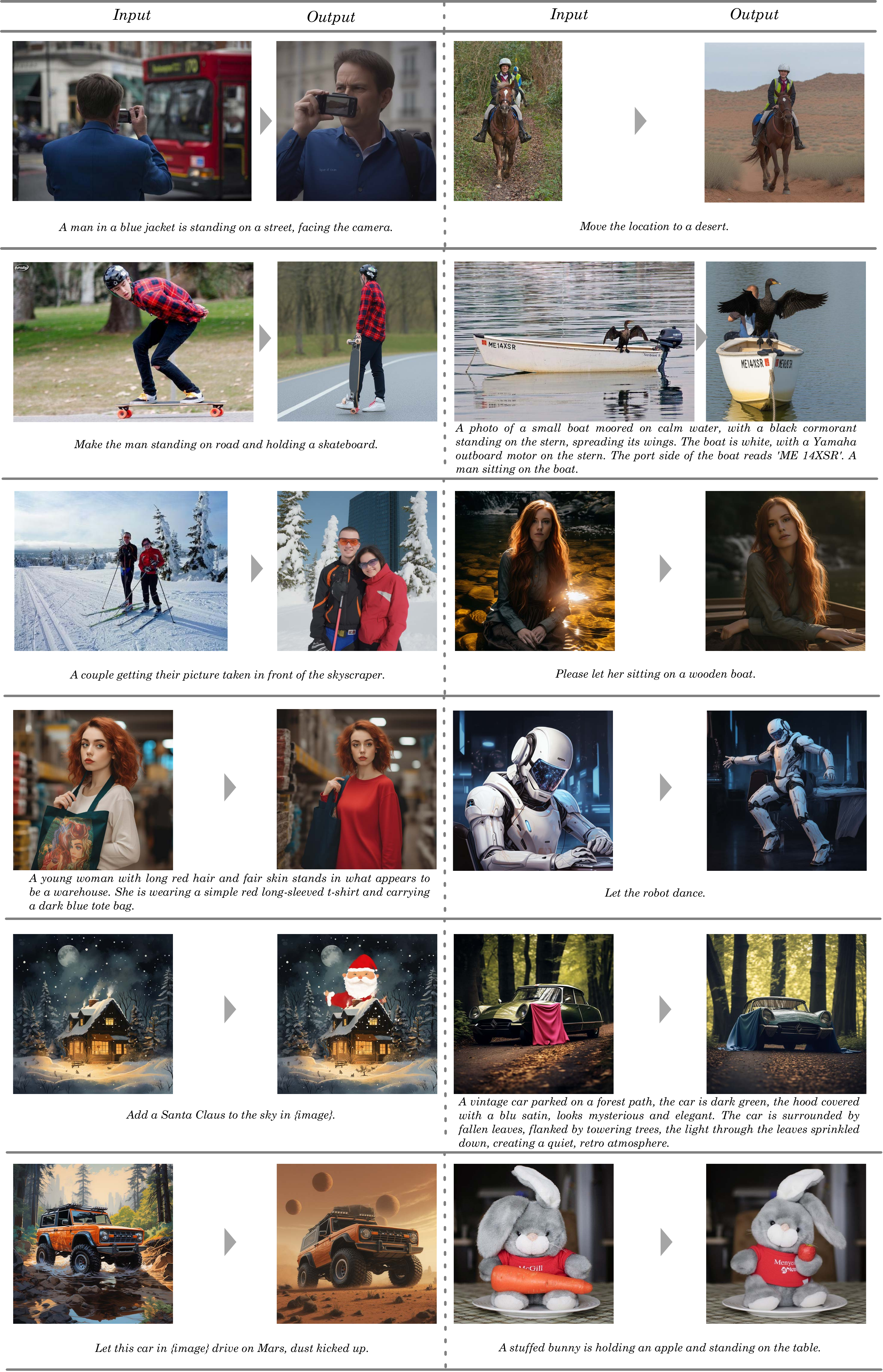}
    \caption{The visualization of the flexible instruction or descriptions}
    \vspace{-15pt}
    \label{fig:general}
\end{figure*}

%% file: figures/local_ref/local_ref.tex
\begin{figure*}[t]
    \scriptsize
    \centering
    \includegraphics[width=\linewidth]{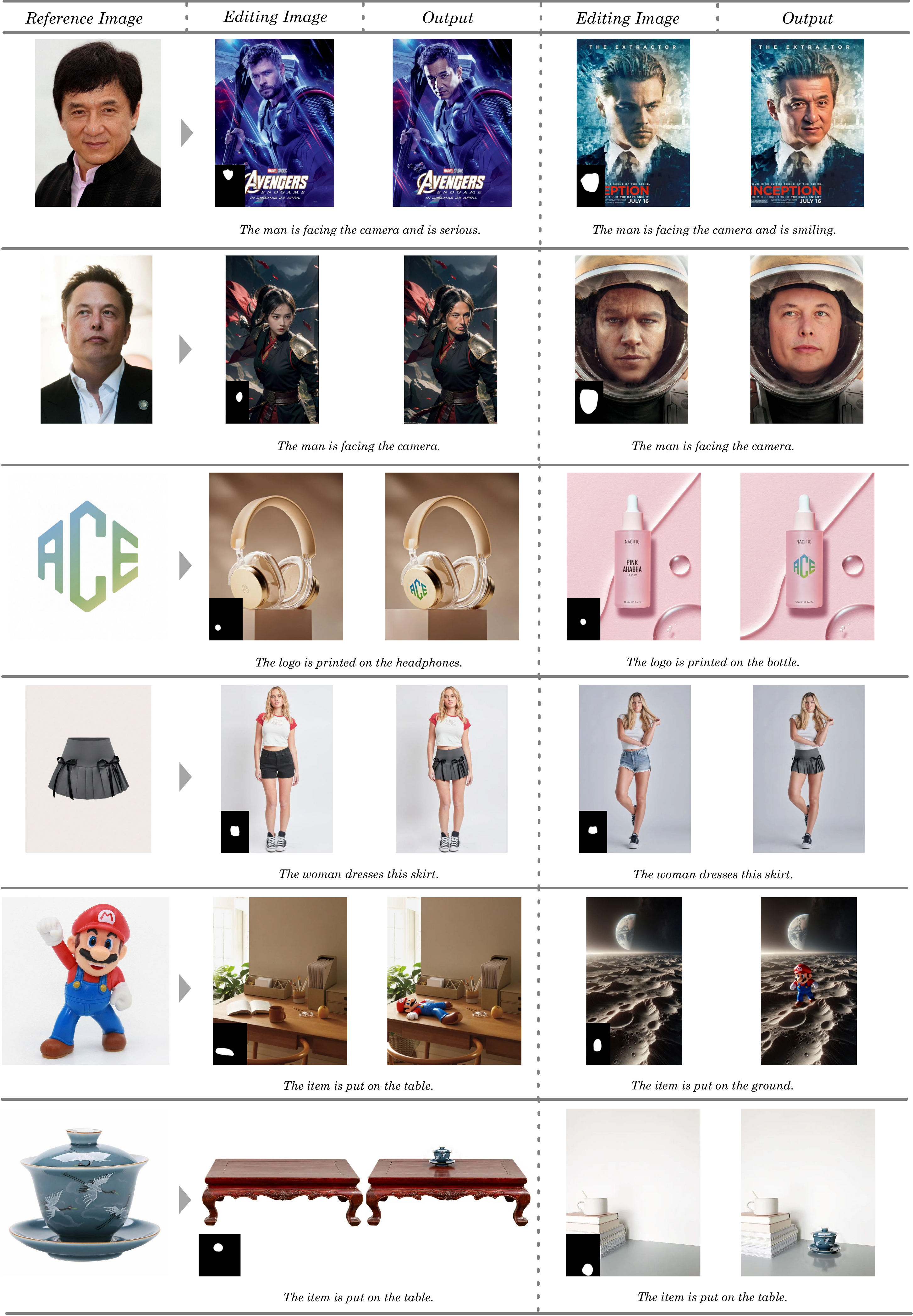}
    \caption{The visualization of local reference editing.}
    \vspace{-15pt}
    \label{fig:local_ref}
\end{figure*}

%% file: sections/5.conclusion.tex
\section{Conclusion}\label{sec:conc}

We propose ACE++, which is an improvement upon ACE, featuring enhanced input paradigms and a two-stage training approach. This allows us to effectively leverage the pre-training capabilities of foundational text-to-image models, significantly reducing the cost and cycle of fine-tuning large-scale multimodal generative models. Additionally, we provide an all-in-one editing and generation model, similar to ACE, as well as vertical models tailored for specific application scenarios, to support downstream innovative applications within the community.